\documentclass{article}

\usepackage[preprint]{neurips_2026}

\usepackage[utf8]{inputenc} 
\usepackage[T1]{fontenc}    
\usepackage{hyperref}       
\usepackage{url}            
\usepackage{booktabs}       
\usepackage{amsfonts}       
\usepackage{nicefrac}       
\usepackage{microtype}      
\usepackage{xcolor}         
\usepackage{amsmath}
\usepackage{graphicx}
\usepackage{multirow}
\usepackage[table]{xcolor}
\usepackage{enumitem}
\usepackage{subcaption}
\usepackage{wrapfig}
\usepackage[most]{tcolorbox}
\newtcblisting{promptbox}{
  enhanced,
  width=\linewidth,     
  colback=gray!5,           
  colframe=black!30,        
  boxrule=0.5pt,           
  arc=3pt,                  
  left=5pt, right=5pt, top=5pt, bottom=5pt, 
  listing only,
  listing options={
    basicstyle=\ttfamily\small\linespread{1.1},
    breaklines=true,      
    breakatwhitespace=false,
    columns=fullflexible,
    keepspaces=true,
    numbers=none
  }
}

\title{Belief Memory:\\ Agent Memory Under Partial Observability}

\author{%
  Junfeng Liao\textsuperscript{\rm 1}, 
  Qizhou Wang\textsuperscript{\rm 2}, 
  Jianing Zhu\textsuperscript{\rm 3}, 
  Bo Du\textsuperscript{\rm 4}, 
  Rui Yan\textsuperscript{\rm 4}, 
  Xiuying Chen\textsuperscript{\rm 1}\thanks{Corresponding author. Email: xiuying.chen@mbzuai.ac.ae} \\
  \\
  \textsuperscript{\rm 1}MBZUAI \quad
  \textsuperscript{\rm 2}RIKEN AIP \quad
  \textsuperscript{\rm 3}UT Austin \quad
  \textsuperscript{\rm 4}Wuhan University \\
}

\begin{document}

\maketitle

\begin{abstract}
LLM agents that operate over long context depend on external memory to accumulate knowledge over time.
However, existing methods typically store each observation as a single deterministic conclusion (e.g., inferring ``API~X failed'' from temporary errors), even though such observations are inherently partial and potentially ambiguous. 
By committing to one conclusion and discarding uncertainty, these methods introduce \emph{self-reinforcing error}: the agent acts on the stored conclusion, never revisits alternatives, and reinforces the conclusion over time.
To address this issue, we propose \emph{BeliefMem}, which shifts the memory paradigm from committing to a single conclusion per observation to retaining multiple candidate conclusions with their probabilities.
Concretely, BeliefMem stores the candidate conclusions as separate memory entries, each carrying a probability that is updated via Noisy-OR rules as new observations arrive.
At retrieval, all candidates surface together with their probabilities, keeping alternatives visible to the agent.
Since each conclusion in memory retains its probability, BeliefMem preserves the uncertainty that the deterministic paradigm discards, enabling the agent to act with high confidence on well-evidenced knowledge while retaining the capacity to update its confidence when new evidence arrives. 
Empirical evaluations on LoCoMo and ALFWorld benchmarks show that, even with limited data, BeliefMem achieves the best average performance, remarkably outperforming well-known baselines.
More broadly, such probabilistic memory produces substantial gains and explores a new direction for agent memory in partially observable environments.
\end{abstract}

\section{Introduction}
\label{sec:intro}
\emph{Large language model} (LLM) agents deployed in long-horizon, multi-session tasks increasingly rely on persistent external memory to accumulate knowledge across interactions~\citep{hu2025memory,du2026memory}. 
\emph{Factual memory} methods store observations about users and environments as structured entries, from natural-language memory streams~\citep{park2023generative} to vector-based extracted facts~\citep{chhikara2025mem0}. 
While these methods record what was observed, \emph{self-improving memory} methods distill actionable lessons from past experience, from natural-language reflections~\citep{shinn2023reflexion,zhao2024expel} to reusable skill libraries~\citep{zhang2026memskill}. 
Despite this diversity, these methods share a common paradigm: every memory entry is stored as a single deterministic conclusion inferred from observations, and every operation over it produces an all-or-nothing outcome.

\begin{figure}
    \centering
    \includegraphics[width=1\linewidth]{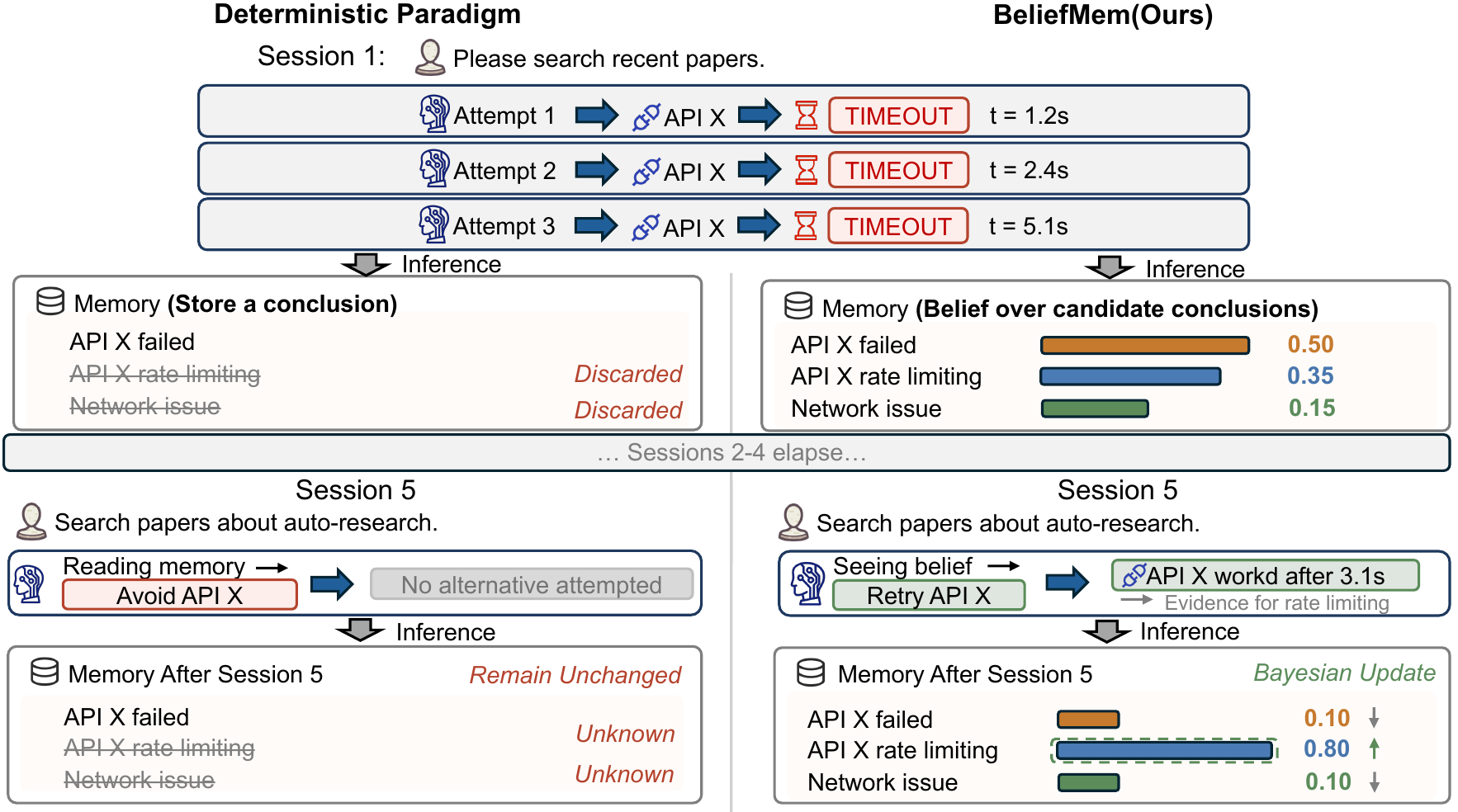}
    \caption{Deterministic memory vs. BeliefMem with an API timeout example. After repeated API X timeouts, the deterministic paradigm stores “API X failed” and avoids it in later sessions, reinforcing the error. In contrast, BeliefMem keeps multiple hypotheses (e.g., failure vs. rate limiting) with probabilities, retries the API, and updates beliefs with new evidence, enabling correction over time.}
    \label{fig:intro}
\end{figure}

This deterministic paradigm results in errors that persist over time. 
Consider an agent that observes repeated API~X timeouts (Figure~\ref{fig:intro}): since each memory entry holds only a single categorical conclusion, the agent stores ``API~X failed'' while the possibility of transient failure (e.g., temporary rate limiting) is permanently discarded. 
Self-improving methods amplify this problem by distilling experience such as ``avoid API~X,'' and even methods that update entries cannot escape, as correcting to ``API~X is operational'' merely replaces one deterministic conclusion with another and the next transient error flips it right back. 
Furthermore, when such flawed conclusions conflict with user instructions (e.g., ``Use API~X to ...''), the agent struggles to act reliably~\citep{hu2025evaluating}. We refer to this issue as \emph{self-reinforcing error}: the agent acts on stored conclusions, generating observations that further evidence them~\citep{shao2025your,lam2026governing}.

Fundamentally,  these agents operate in a \emph{partially observable Markov decision process} (POMDP): they never directly access the true state of the world but only receive partial, noisy observations such as user messages and tool outputs~\citep{kaelbling1998planning}. For instance, whether API~X is permanently down or temporarily rate-limited is a hidden state that must be inferred from observations. Yet existing deterministic memory methods equate each observation with ground truth, leaving alternative hypotheses unrepresented and allowing self-reinforcing errors to persist across sessions (Figure~\ref{fig:intro}).

To bridge this gap, we propose \textit{BeliefMem}, which fundamentally shifts the memory paradigm from storing deterministic conclusions to maintaining an attribute-level belief representation over the environment.
Specifically, BeliefMem maintains active candidate conclusions for each piece of stored knowledge, assigning each conclusion a probability updated via noisy-OR evidence merge as new observations arrive.
At retrieval, the candidate conclusions of each latent state surface with their probabilities, keeping competing hypotheses visible to the agent instead of reducing them to a single deterministic conclusion.
This combination of belief-aware memory storage and probability-aware retrieval directly mitigates self-reinforcing error at its root: the alternative conclusions that the deterministic paradigm discards during the storage phase are now preserved and accessible to the agent. 
For example, in Figure~\ref{fig:intro}, repeated timeouts on API~X keep candidate conclusions viable alongside permanent failure.
Therefore, the agent can revisit previously unfavorable actions in the future, and each new observation incrementally refines the probability assignment of each conclusion, strengthening well-supported conclusions and downweighting those with weak evidence.

To evaluate BeliefMem, we conduct experiments on both LoCoMo~\citep{maharana2024evaluating} and ALFWorld~\citep{shridhar2020alfworld} benchmarks, from long-term conversation to embodied agent interaction settings. Empirical evaluations show that our method achieves the best average performance on both benchmarks, outperforming existing memory methods, even with limited memory corpus size. 
Furthermore, ablation studies and adversarial experiments confirm the effectiveness of BeliefMem in preserving uncertainty and refining memories.
More broadly, these results demonstrate that replacing deterministic memory entries with probabilistic belief representation yields promising gains, exploring a new direction for agent memory paradigm in partially observable environments.

\section{Related Work}

\subsection{Factual and RL-Based Memory}
Factual and RL-based memory methods follow the deterministic paradigm, reducing each observation's candidate conclusions to a single categorical one and discarding the alternatives. 
Within this shared paradigm, early factual memory methods differ mainly in how they organize and access stored entries. Generative Agents~\citep{park2023generative} maintains a natural language memory stream and retrieves memories with various signals, whereas MemGPT~\citep{packer2023memgpt} manages memories across context, recall, and storage through virtual context management. 
Subsequent work further improves extraction, organization, and retrieval without changing the underlying representation: Mem0~\citep{chhikara2025mem0} dynamically extracts and consolidates salient facts for vector-based retrieval, and A-MEM~\citep{xu2025mem} organizes memories as structured notes with indexing and linking.
Other work enriches the storage structure itself, with MemoryBank~\citep{zhong2024memorybank} updating retrieval strength with a forgetting curve, Zep~\citep{rasmussen2025zep} preserving evolving information in a temporal knowledge graph, and MemOS~\citep{li2025memos} unifying heterogeneous memory blocks within a single system. 
Meanwhile, RL-based memory methods replace this hand-crafted memory management with learnable policies to add/update/delete entries, including Memory-R1~\citep{yan2025memory}, MEM1~\citep{zhou2025mem1}, Agentic Memory~\citep{yu2026agentic}, and MemRL~\citep{zhang2026memrl}.
Across these studies, the main differences lie in storage management and retrieval strategy, not in memory representation, where each memory entry generally still records only one categorical conclusion inferred from noisy and ambiguous observations.

\subsection{Self-improving Memory}
Beyond recording factual observations, self-improving memory methods store actionable lessons distilled from past experience to instruct the agent's subsequent actions.
There are several studies that summarize raw experience into verbal lessons, such as Generative Agents~\citep{park2023generative} summarizing interaction history as reflective memory, Reflexion~\citep{shinn2023reflexion} generating self-corrective guidance from failed experiences, and ExpeL~\citep{zhao2024expel} aggregating recurring patterns across trajectories into reusable insights.
Beyond verbal lessons, concurrent work records feasible actions in growing skill libraries. Voyager~\citep{wang2023voyager} expands the library through an automatic curriculum as the agent explores new environments, and MemSkill~\citep{zhang2026memskill} constructs a set of skills that transfer reusable knowledge across related problems. Despite shifting from factual observations to distilled experience, these methods retain the same deterministic paradigm, storing each lesson as a single categorical entry while ignoring uncertainty in observations.

\subsection{Belief State under Partial Observability}
In the standard POMDP, uncertainty under partial observability is represented by a belief state, a probability distribution over hidden states conditioned on the observation history~\citep{kaelbling1998planning}. 
Recent work views LLM agents as operating under partial observability and uses belief based representations for action selection and coordination~\citep{lidayan2025abbel,jiang2026pabu,wang2025cobel}. 
Additionally, Belief Engine~\citep{yang2026belief} externalizes and updates beliefs in a specific multi-agent debate setting, and empirical work shows that the mismatch between agent's beliefs and the true states of the environment can result in unreliable opinions and actions~\citep{geng2025accumulating}.
However, existing memory systems still ignore the key implication of such partial observability, namely that an agent's observations provide only partial evidence about hidden states (e.g., user preference) rather than direct access to the true states.
As a result, memory is represented as deterministic conclusions inferred from noisy observations, collapsing their uncertainty into a single ground truth.
This motivates a memory representation that preserves such uncertainty instead of storing each memory entry as ground truth.

\section{Methodology}

\subsection{Problem Formulation}
\label{subsec:problem}
\textbf{POMDP (Partially Observable Markov Decision Process) setting.}
We consider an agent interacting with partially observable environments. At decision time $t$, the agent has access to an observation $o_t \in \mathcal{O}$ and selects an action $a_t \in \mathcal{A}$. Let $s_t \in \mathcal{S}$ denote the latent environment state at time $t$, the environment transitions according to $s_{t+1} \sim T(\cdot \mid s_t, a_t)$~\citep{kaelbling1998planning}.  Bayes-optimal action selection depends on the belief state, i.e., the posterior distribution over latent states induced by the interaction history. 
Defining $\eta_t := (o_{1:t}, a_{1:t-1})$, we write:
\begin{equation}
    b_t(s) := \Pr(s_t = s \mid \eta_t), \qquad b_t \in \Delta(\mathcal S), \qquad a_t \sim \pi(\cdot \mid b_t).
\end{equation}
Therefore, $b_t$ is a sufficient statistic of the action-observation history for action selection. 

\textbf{External Memory as Belief Approximation.}
Existing memory methods can be viewed as approximating $b_t$ through an external memory module $M_t$, which compresses task-relevant information from past interactions into a retrievable structure. At $t$, the agent queries $M_t$ with the current observation $o_t$ to obtain the memory context:
\begin{equation}
\label{eq:memory-read}
z_t = \mathrm{Read}(M_t, o_t),
\end{equation}
and selects an action conditioned on both the observation and the retrieved context: $a_t \sim \pi(\cdot \mid o_t, z_t)$. After executing $a_t$ and observing $o_{t+1}$, the memory is updated as:
\begin{equation}
\label{eq:memupdate}
M_{t+1} = \mathrm{Update}(M_t, o_t, o_{t+1}),
\end{equation}
where $\mathrm{Update}$ encompasses memory writing and management operations~\citep{xu2025mem,yan2025memory}. In this way, $M_t$ serves as a tractable approximation of the belief state, supporting future decisions without maintaining the inaccessible full posterior.

\subsection{Motivation}
\label{subsec:motivation}

\textbf{The Deterministic Bottleneck.}
However, in practice, many existing memory methods store point estimates of latent attributes relevant to the task, i.e., a deterministic conclusion of each attribute inferred from observations, thus discarding uncertainty that would be retained in a representation of complete belief $b_t(s)$.
Let $c$ denote a task-relevant attribute of the latent state (e.g., user preference, tool status, or
object-location relation), and let $\mathcal{H}(c) = \{h_1^{(c)}, \dots, h_{M_c}^{(c)}\}$ denote a set of mutually exclusive and collectively exhaustive hypotheses representing the possible conclusions of $c$.
A reliable memory would maintain, for each $c$, a local posterior:
\begin{equation}
    b_t^{(c)}(h):=\Pr(s_t\in h\mid o_{1:t},a_{1:t-1})=\textstyle \sum_{s\in h} b_t(s), \qquad h\in\mathcal H(c).
\end{equation}
However, in the deterministic memory paradigm, the write operation stores only a single conclusion $\hat h_t(c)$ rather than the full local posterior:
\begin{equation}
\label{eq:deterministic-M}
    M_t=\{(c,\hat h_t(c)):c\in\mathcal C_t\}, \qquad \hat h_t(c)\in\mathcal H(c),
\end{equation}
where $\mathcal{C}_t$ denotes all attributes preserved in $M_t$. Fundamentally, this corresponds to writing the most probable attribute-level hypothesis, $\hat h_t(c)\in\operatorname{arg max}_{h\in\mathcal H(c)} b_t^{(c)}(h)$,
while discarding the remaining alternatives and their associated probabilities. Therefore, $M_t$ in current methods is a collection of attribute-level point estimates rather than a probabilistic approximation to the full belief state $b_t$, and the discarded uncertainty is no longer available for subsequent retrieval or update.

\textbf{Self-Reinforcing Error.}
This point estimate can induce self-reinforcing error.
Suppose the retrieved memory $z_t$ in Eq.~\ref{eq:memory-read} exposes a stored conclusion $(c,\hat h_t(c)) \in M_t$. The agent selects $a_t$ conditioned on $o_t$ and $z_t$, and the resulting transition $(o_t,a_t,o_{t+1})$ is written back to memory via Eq.~\ref{eq:memupdate}. Since memory retains no posterior support for alternative hypotheses in $\mathcal{H}(c) \setminus \{\hat h_t(c)\}$, the agent is unlikely to select actions that would test these alternatives. If $\hat h_t(c)$ is incorrect or prematurely consolidated, the agent instead collects further evidence consistent with the flawed conclusion, reinforcing it over time. 
For example, if memory stores ``API~X failed,'' the agent becomes less likely to retry the API, thereby missing observations that could contradict the stored memory entry. 
Once uncertainty is collapsed to a point estimate, posterior support for discarded alternatives cannot be reconstructed from memory alone and must be re-established through entirely new evidence. This motivates a memory paradigm that retains a belief over the uncertainty rather than collapsing it to a point estimate.

\subsection{Belief Memory}
\label{subsec:belief-memory}

\textbf{Belief-based Memory Formulation.}
To bridge this gap, we propose BeliefMem, which replaces the deterministic paradigm with an attribute-level belief representation that approximates the belief state $ b_{t}^{(c)}$.
We first introduce the ideal representation for each memory entry:
\begin{equation}
\label{eq:belief-entry}
\big(c,\; b_{t}^{(c)}\big), \qquad
b_{t}^{(c)} : \mathcal{H}(c) \to [0,1],  \quad \text{s.t.} \textstyle \sum_{h \in \mathcal{H}(c)} b_{t}^{(c)}(h) = 1,
\end{equation}
where $b_{t}^{(c)}$ denotes the distribution of all possible conclusions for attribute $c$ at time $t$.
Therefore, the ideal representation of memory in BeliefMem is:
\begin{equation}
\label{eq:belief-M}
M_t \;=\; \big\{\big(c,\, b_{t}^{(c)}\big) : c \in \mathcal{C}_t\big\},
\end{equation}
which replaces the deterministic collection of point estimates in Eq.~\ref{eq:deterministic-M} with a belief state which can represent the uncertainty of each attribute in the environment.

In practice, this idealized representation is not directly feasible, because the conclusion space associated with an attribute is open-ended or dynamically expanding, making exact posterior maintenance over the full set impractical. 
This leads to two practical challenges:
i) In open-ended settings, $\mathcal{H}(c)$ is not fixed and may expand online as new candidate conclusions are generated. A fully normalized distribution over all candidate conclusions is therefore difficult to define.
ii) Even under a fixed set of possible conclusions, updating the distribution for all candidates after each new observation is computationally expensive. Therefore, classical POMDP methods rely on approximate belief representations, such as representative belief points, rather than exact update across all possible states.

\begin{figure}
    \centering
    \includegraphics[width=1\linewidth]{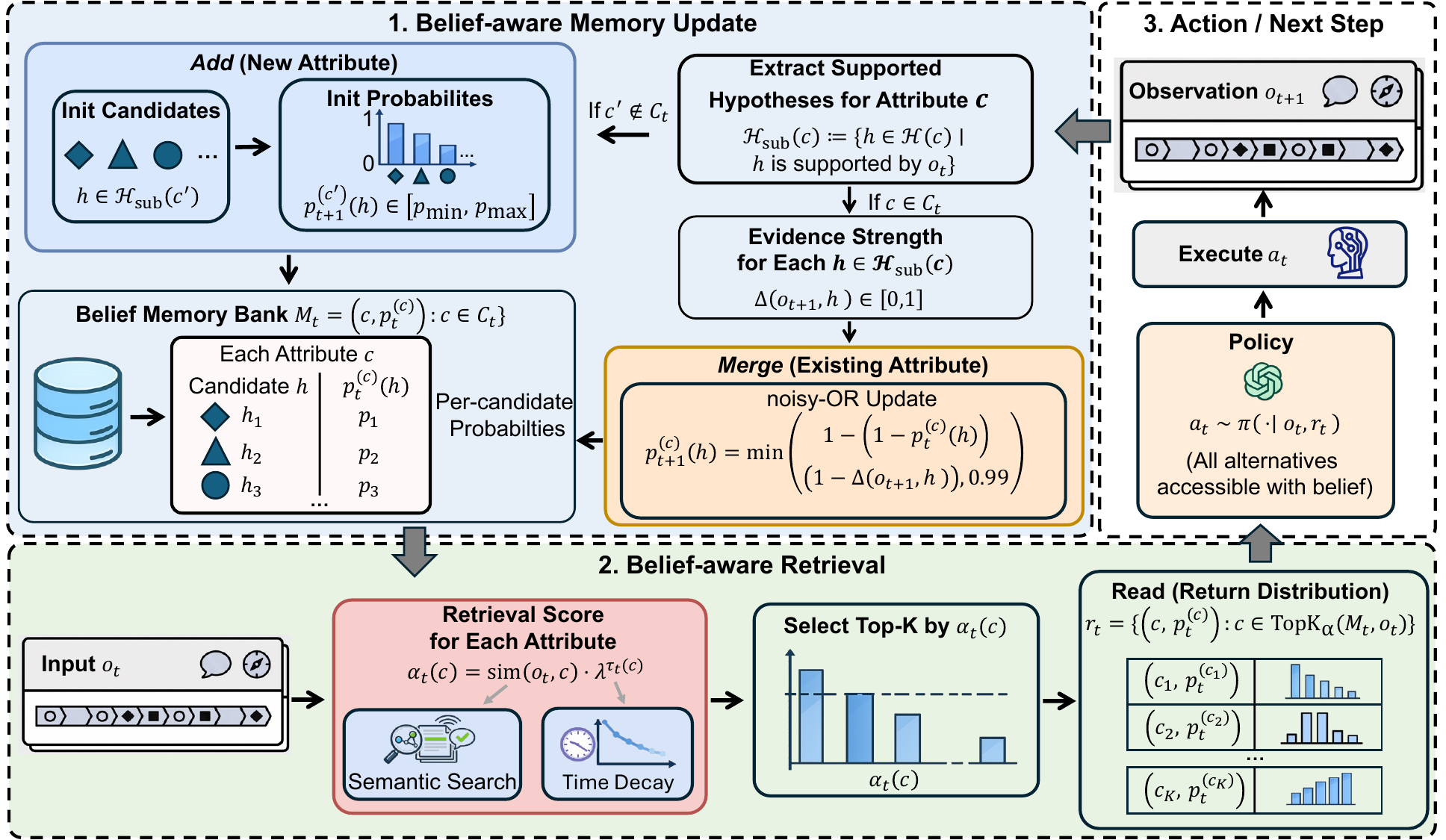}
    \caption{Overview of BeliefMem. i) Upon receiving an observation, BeliefMem updates memories via Add (initializing candidates for new attributes) or Merge (incorporating new evidence via noisy-OR update). ii) Retrieval scores entries by semantic similarity and temporal decay, returning a full belief rather than a single conclusion. iii) The agent acts conditioned on both the current observation and the retrieved belief, keeping all alternative hypotheses visible at decision time.}
    \label{fig:beliefmem}
\end{figure}

\textbf{Belief Update in Memory.}
\label{subsec:update-rules}
To overcome these challenges, in this work, BeliefMem leverages two coupled ways to practically approximate Eq.~\ref{eq:belief-entry}. 
First, for each attribute $c$, BeliefMem only stores candidates that previous observations have actually evidenced, so preserved conclusions grow with evidence rather than with $|\mathcal{H}(c)|$, and unseen candidates incur no storage or update cost. 
Specifically, for each observation $o_t$, the agent identifies the supported hypotheses to form the subset $\mathcal{H}_{\rm sub}(c) := \{h \in \mathcal{H}(c) \mid h \text{ is supported by } o_t\}$. 
The agent then assigns to each $h \in \mathcal{H}_{\rm sub}(c)$ a probability $p_{t}^{(c)} (h) \in [0,1]$ measuring how strongly ``$h$ is true'' under $o_t$, and stores each resulting (attribute $c$, candidate $h$, probability $p_{t}^{(c)} (h)$) into memory bank. 
Secondly, BeliefMem keeps per-candidate probability $\{p_{t}^{(c)}(h)\}$ instead of a normalized joint posterior over $\mathcal{H}_{\rm sub}(c)$ to prevent the magnitude of a supported $h$ from being affected by the number of alternative candidates sharing the same $c$. 
Thus, these values are evidence-based probabilities rather than posterior probabilities over mutually exclusive hypotheses.
While stored independently, these probabilities are updated jointly whenever new evidence for $c$ is observed.

Building upon these principles, BeliefMem dynamically maintains $M_t$ via the following operations:

\textit{\textbf{Add}} works when the agent reports a new attribute $c' \notin \mathcal{C}_t$ and stores a new entry in $M_t$,
\begin{equation}
\label{eq:add}
\big( c',\;h,\; p_{t+1}^{(c')}(h)\big), \qquad p_{t+1}^{(c')}(h) \in [p_{\rm min},\, p_{\rm max}],
\end{equation}
where $h \in \mathcal{H}_{\rm sub}(c')$ is supported by the current observation $o_{t+1}$. $p_{\rm min}$ and $p_{\rm max}$ constrain the probability of each new conclusion. The details of extracting attributes are provided in Appendix~\ref{app:add}.

\textit{\textbf{Merge}} activates when the new observation supports an attribute $c \in \mathcal{C}_t$ already present in $M_t$.
For any candidate conclusion $h$, if the observation provides supporting evidence, its belief is updated via noisy-OR evidence merge:
\begin{equation}
\label{eq:merge}
p_{t+1}^{(c)}(h) = 
\min\!\Bigl(1 - \bigl(1 - p_t^{(c)}(h)\bigr)\bigl(1 - \Delta(o_{t+1}, h)\bigr),\; 0.99\Bigr),
\end{equation}
where $\Delta(o_{t+1}, h) \in [0,1]$ quantifies the strength of evidence provided by $o_{t+1}$ for the stored conclusion $h$ (details are provided in Appendix~\ref{app:add}). The upper bound of 0.99 prevents any candidate from being stored with certainty. After \emph{Merge}, BeliefMem archives the old version $p_t^{(c)}$ for later retrieval. Additionally, if the observation supports a competing candidate for the same attribute $c$, its probability would be reduced to $0.25$. Details are presented in Appendix~\ref{app:contra}.

\textbf{Belief-aware Retrieval.}
Belief update alone is insufficient if retrieval discards the uncertainty 
that storage has carefully preserved. 
To close this gap, retrieval is redefined as an operation conditioned on the stored belief rather than on a single chosen conclusion.
Specifically, given an observation $o_t$, the retrieval score of each entry is:
\begin{equation}
\label{eq:decay}
\alpha_t(c) \;=\; \mathrm{sim}(o_t,\, c) \cdot \lambda^{\tau_t(c)}, \qquad \lambda \in (0,1],
\end{equation}
where $\mathrm{sim}(\cdot) \in \mathbb{R}_{\geq 0}$ measures the relevance of $c$ to $o_t$ through semantic similarity. The specific choice of $\mathrm{sim}$ is shown in Appendix~\ref{app:hyper}. $\lambda \in [0,1]$ is the decay rate to control temporal importance during retrieval.
$\tau_t(c) \in \mathbb{N}$ denotes the staleness of entry $c$ (i.e., the time elapsed since its last update). It increases by one at each time step, unless the corresponding entry $c$ is updated by \emph{Add} or \emph{Merge}, in which case it resets to $0$. Thus, an entry's retrieval priority decays with staleness, and its underlying probability mass remains unchanged. $\mathrm{Read}$ then selects the top-$K$ entries by $\alpha_t(c)$ and returns
\begin{equation}
\label{eq:belief-read}
r_t \;=\; \big\{\big(c,\, p_t^{(c)}\big) : c \in \mathrm{TopK}_{\alpha}(M_t, o_t)\big\},
\end{equation}
so that each retrieved attribute has its candidate probabilities over $\mathcal{H}_{\rm sub}(c)$. The agent then selects an action as $a_t \sim \pi(\cdot \mid o_t, r_t)$, and every alternative conclusion in $\mathcal{H}_{\rm sub}(c)$ is now accessible to the agent with its confidence, rather than being erased at storage time in the deterministic paradigm.

Overall, BeliefMem mitigates self-reinforcing error through two coupled principles. Specifically, it preserves memory as an approximated belief representation and returns the candidate beliefs at retrieval so that alternative hypotheses remain visible to the agent at decision time.

\section{Experiments}
\begin{table}[t]
\centering
\caption{LoCoMo results across four categories under GPT-4o-mini and GPT-4o backbones. Each cell reports F1 / BLEU-1. Best and second numbers per column are in \textbf{bold} and \underline{underline}, respectively.}
\label{tab:locomo}
\footnotesize
\resizebox{\linewidth}{!}{ 
\begin{tabular}{llccccc}
\toprule
Backbone & Method & Multi-Hop & Temporal & Open Domain & Single-Hop & Avg. \\
\midrule
\multirow{9}{*}{GPT-4o-mini}
 & LoCoMo     & 25.02 / 19.75 & 18.41 / 14.77 & 12.04 / 11.16 & 40.36 / 29.05 & 23.96 / 18.68 \\
 & ReadAgent  &  9.15 /\phantom{0}6.48 & 12.60 /\phantom{0}8.87 &  5.31 /\phantom{0}5.12 &  9.67 /\phantom{0}7.66 & \phantom{0}9.18 /\phantom{0}7.03 \\
 & MemoryBank &  5.00 /\phantom{0}4.77 &  9.68 /\phantom{0}6.99 &  5.56 /\phantom{0}5.94 &  6.61 /\phantom{0}5.16 & \phantom{0}6.71 /\phantom{0}5.72 \\
 & MemGPT       & 26.65 / 17.72 & 25.52 / 19.44 &  9.15 /\phantom{0}7.44 & 41.04 / 34.34 & 25.59 / 19.74 \\
 & A-MEM$^{*}$              & 27.02 / 20.09 & 45.85 / 36.67 & 12.14 / 12.00 & 44.65 / 37.06 & 32.42 / 26.46 \\
 & A-MEM                          & 27.08 / 20.46 & 29.14 / 24.08 & 16.60 / 13.80 & 40.70 / 32.63 & 28.38 / 22.74 \\
 & Mem0$^{*}$         & \underline{38.72} / \underline{27.13} & \underline{48.93} / \underline{40.51} & \underline{28.64} / \underline{21.58} & \underline{47.65} / \underline{38.72} & \underline{40.99} / \underline{31.99} \\
 & Mem0                            & 36.83 / 26.50 & 34.52 / 26.38 & 22.57 / 16.54 & 46.89 / 37.63 & 35.20 / 26.76 \\
 \rowcolor{gray!20} \cellcolor{white} & \textbf{BeliefMem (ours)}             & \textbf{40.51} / \textbf{32.24} & \textbf{51.88} / \textbf{45.78} & \textbf{28.73} / \textbf{25.12} & \textbf{48.41} / \textbf{42.07} & \textbf{42.38} / \textbf{36.30} \\
\midrule
\multirow{8}{*}{GPT-4o}
 & LoCoMo    & 28.00 / 18.47 &  9.09 /\phantom{0}5.78 & 16.47 / 14.80 & \textbf{61.56} / \textbf{54.19} & 28.78 / 23.31 \\
 & ReadAgent  & 14.61 /\phantom{0}9.95 &  4.16 /\phantom{0}3.19 &  8.84 /\phantom{0}8.37 & 12.46 / 10.29 & 10.02 /\phantom{0}7.95 \\
 & MemoryBank &  6.49 /\phantom{0}4.69 &  2.47 /\phantom{0}2.43 &  6.43 /\phantom{0}5.30 &  8.26 /\phantom{0}7.10 & \phantom{0}5.91 /\phantom{0}4.88 \\
 & MemGPT      & 30.36 / 22.83 & 17.29 / 13.18 & 12.24 / 11.87 & \underline{60.18} / \underline{53.35} & 30.02 / 25.31 \\
 & A-MEM$^{*}$               & 32.86 / 23.76 & 39.41 / 31.23 & 17.10 / 15.84 & 48.43 / 42.97 & 34.45 / 28.45 \\
 & A-MEM                           & 31.66 / 23.34 & 41.11 / \underline{34.72} & 17.45 / 15.58 & 47.04 / 41.02 & 34.32 / 28.67 \\
 & Mem0                           & \textbf{42.57} / \underline{30.92} & \underline{44.55} / 32.60 & \underline{23.04} / \underline{17.62} & 48.49 / 37.00 & \underline{39.66} / \underline{29.54} \\
 \rowcolor{gray!20} \cellcolor{white} & \textbf{BeliefMem (ours)}             & \underline{41.24} / \textbf{35.68} & \textbf{50.81} / \textbf{45.05} & \textbf{27.60} / \textbf{23.78} & 51.81 / 43.80 & \textbf{42.87} / \textbf{37.08} \\
\bottomrule
\end{tabular}
}
\end{table}

\textbf{Benchmarks.} We conduct experiments on two benchmarks to evaluate long-term memory capabilities of BeliefMem in both long-term conversation and embodied agent interaction settings: i) \emph{LoCoMo}~\citep{maharana2024evaluating}, a long-term conversational memory benchmark whose dialogues contain roughly 9,000 tokens on average and up to 35 sessions, stressing multi-session retrieval and temporal reasoning. Following this, we evaluate along four question categories: \emph{single-hop}, which asks the model to extract a specific fact from a single session; \emph{multi-hop}, which requires composing information scattered across multiple sessions; \emph{temporal reasoning}, which evaluates ordering and duration of events along the dialogue timeline; and \emph{open-domain}, which demands combining the contextual history with external commonsense knowledge. We report F1 for token-level precision and recall, and BLEU-1 for lexical overlap against ground-truth answers.
ii) \emph{ALFWorld}~\citep{shridhar2020alfworld}, a text-based embodied benchmark whose tasks cover six household goal categories. The evaluation is split into an in-distribution \emph{Seen} set and an out-of-distribution \emph{Unseen} set whose room layouts and object instances are held out from training, so the latter directly probes memory transfer rather than pattern memorization. We report success rate (SR), the fraction of tasks whose goal condition is satisfied within a 50-step horizon, and the average step used on solved episodes, following~\citet{zhang2026memskill}. 
More details on the evaluations are provided in the Appendix~\ref{app:alfworld}.

\textbf{Baselines.} On LoCoMo, we compare BeliefMem with six well-known memory methods: LoCoMo baseline from~\citet{maharana2024evaluating}, ReadAgent~\citep{lee2024human}, MemoryBank~\citep{zhong2024memorybank}, MemGPT~\citep{packer2023memgpt}, A-MEM~\citep{xu2025mem}, and Mem0~\citep{chhikara2025mem0}. $^{*}$ denotes official reported performance. 
On ALFWorld, we extend these baselines with LangMem~\citep{LangMem} and MemoryOS~\citep{kang2025memory}, and include No-Memory that chooses actions directly from the current observation to show the contribution of memory.

\textbf{Implementation Details.} On LoCoMo, we use text-embedding-3-small for embedding, and utilize GPT-4o and GPT-4o-mini~\citep{hurst2024gpt} as base models and Qwen3-Next-80B-A3B-Instruct~\citep{yang2025qwen3} as the base model for ALFWorld. All baselines are run with their released configurations. For BeliefMem, $p_{\rm min}$ and $p_{\rm max}$ are shared across benchmarks, while the decay rate $\lambda$ is set per benchmark. The hyperparameter configurations of all methods are listed in Appendix~\ref{app:hyper}.

\subsection{Main results}
\label{sub:mainresult}

\textbf{Effectiveness in long conversational scenarios}. As demonstrated in Table \ref{tab:locomo}, BeliefMem achieves the highest average performance across both base models, producing substantial improvements in multi-hop and temporal reasoning tasks. These specific tasks rigorously test an agent's ability to resolve observation conflicts and aggregate evidence over interactions.
The effectiveness of BeliefMem arises from its dynamic belief update mechanism, which continuously refines and retains essential historical context while mitigating memory degradation. 
Furthermore, by archiving prior memory states with explicit temporal metadata, BeliefMem supports more precise retrieval of former environmental states, directly facilitating its superior temporal reasoning.

\textbf{Superiority in embodied interactive scenarios.} As detailed in Table~\ref{tab:alfworld}, BeliefMem consistently outperforms all baselines across seen and unseen tasks. 
Specifically, BeliefMem outperforms the second-best method (ReadAgent) by 11\%, and exceeds the average of the remaining baselines by 99\% overall.
This advantage grows to 12.4\% over the second-best baseline in unseen (out-of-distribution) scenarios, demonstrating BeliefMem's robust generalizability in real agent scenario with memory.
Crucially, BeliefMem achieves this superior performance using only half of the standard memory corpus. 
In Section~\ref{sub:further}, we provide a detailed analysis of this remarkable data efficiency, showing that only 16.67\% of the memory corpus is sufficient to outperform 5 out of 6 baselines.

\begin{table}[t]
\centering
\caption{ALFWorld results with the Qwen3-Next-80B-A3B-Instruct, on the in-distribution (Seen) split and the out-of-distribution (Unseen) split. SR(\%): Success rate ($\uparrow$); \#Steps: Average steps on solved episodes ($\downarrow$). $\Delta$ indicates the difference relative to the best result in each column. BeliefMem*: 50\% memory corpus used. BeliefMem: full memory corpus used.}
\label{tab:alfworld}
\footnotesize
\setlength{\tabcolsep}{2.5pt}
\resizebox{\textwidth}{!}{
\begin{tabular}{l cccccccccccc}
\toprule
\multirow{3}{*}{Method} & \multicolumn{4}{c}{ALF-Seen} & \multicolumn{4}{c}{ALF-Unseen} & \multicolumn{4}{c}{Avg.} \\
\cmidrule(lr){2-5} \cmidrule(lr){6-9} \cmidrule(lr){10-13}
 & \multicolumn{2}{c}{SR $\uparrow$} & \multicolumn{2}{c}{\#Steps $\downarrow$} & \multicolumn{2}{c}{SR $\uparrow$} & \multicolumn{2}{c}{\#Steps $\downarrow$} & \multicolumn{2}{c}{SR $\uparrow$} & \multicolumn{2}{c}{\#Steps $\downarrow$} \\
\cmidrule(lr){2-3} \cmidrule(lr){4-5} \cmidrule(lr){6-7} \cmidrule(lr){8-9} \cmidrule(lr){10-11} \cmidrule(lr){12-13}
 & Val. & $\Delta$ & Val. & $\Delta$ & Val. & $\Delta$ & Val. & $\Delta$ & Val. & $\Delta$ & Val. & $\Delta$ \\
\midrule
No-Memory               & 18.57 & \textcolor{gray}{-45.00} & 42.48 & \textcolor{gray}{+14.99} & 26.12 & \textcolor{gray}{-35.07} & 39.35 & \textcolor{gray}{+11.94} & 22.35 & \textcolor{gray}{-37.53} & 40.92 & \textcolor{gray}{+13.27} \\
ReadAgent               & 53.57 & \textcolor{gray}{-10.00} & 27.88 & \textcolor{gray}{+0.39} & \underline{54.48} & \textcolor{gray}{-6.71} & 27.41 & - & 54.03 & \textcolor{gray}{-5.85} & 27.65 & - \\
MemoryBank              & 37.86 & \textcolor{gray}{-25.71} & 35.15 & \textcolor{gray}{+7.66} & 38.06 & \textcolor{gray}{-23.13} & 34.99 & \textcolor{gray}{+7.58} & 37.96 & \textcolor{gray}{-21.92} & 35.07 & \textcolor{gray}{+7.42} \\
A-MEM                   & 25.00 & \textcolor{gray}{-38.57} & 40.28 & \textcolor{gray}{+12.79} & 29.10 & \textcolor{gray}{-32.09} & 39.04 & \textcolor{gray}{+11.63} & 27.10 & \textcolor{gray}{-32.78} & 39.66 & \textcolor{gray}{+12.01} \\
Mem0                    & 38.57 & \textcolor{gray}{-25.00} & 33.64 & \textcolor{gray}{+6.15} & 41.04 & \textcolor{gray}{-20.15} & 33.16 & \textcolor{gray}{+5.75} & 39.81 & \textcolor{gray}{-20.07} & 33.40 & \textcolor{gray}{+5.75} \\
LangMem                 & 37.14 & \textcolor{gray}{-26.43} & 34.42 & \textcolor{gray}{+6.93} & 31.34 & \textcolor{gray}{-29.85} & 37.17 & \textcolor{gray}{+9.76} & 34.24 & \textcolor{gray}{-25.64} & 35.80 & \textcolor{gray}{+8.15} \\
MemoryOS                & 19.29 & \textcolor{gray}{-44.28} & 42.43 & \textcolor{gray}{+14.94} & 18.66 & \textcolor{gray}{-42.53} & 42.95 & \textcolor{gray}{+15.54} & 18.98 & \textcolor{gray}{-40.90} & 42.69 & \textcolor{gray}{+15.04} \\
\rowcolor{gray!20} \textbf{BeliefMem* (ours)} & \underline{58.57} & \textcolor{gray}{-5.00} & 29.77 & \textcolor{gray}{+2.28} & \textbf{61.19} & - & 29.34 & \textcolor{gray}{+1.93} & \textbf{59.88} & - & 29.56 & \textcolor{gray}{+1.91} \\
\rowcolor{gray!20} \textbf{BeliefMem (ours)}  & \textbf{63.57} & - & 27.49 & - & 53.75 & \textcolor{gray}{-7.44} & 30.49 & \textcolor{gray}{+3.08} & \underline{58.66} & \textcolor{gray}{-1.22} & 28.99 & \textcolor{gray}{+1.34} \\
\bottomrule
\end{tabular}
}
\end{table}

\begin{wraptable}{R}{0.57\textwidth}
    \centering
    \vspace{-5mm}
    \caption{Results of ablation studies on LoCoMo (GPT-4o-mini) and ALFWorld (Qwen3-Next-80B-A3B-Instruct). w/o memory: without belief-based memory; w/o retrieval: without belief-aware retrieval.}
    \label{tab:ablation}  
    \begin{tabular}{lcccc}
        \toprule
        & \multicolumn{2}{c}{LoCoMo} & \multicolumn{2}{c}{ALFWorld} \\
        \cmidrule(lr){2-3} \cmidrule(lr){4-5}
        & F1$\uparrow$ & BLEU$\uparrow$ & SR$\uparrow$ & \#Steps $\downarrow$ \\
        \midrule
        w/o memory    & 22.58 & 16.92 & 28.71 & 38.87 \\
        w/o retrieval & 28.50  & 19.02 & 51.77 & 32.57 \\
        w/o \emph{Add}             & 14.48 & 10.14 & 22.58 & 41.53 \\
        w/o \emph{Merge}           & 20.38 & 13.31  & 40.81 & 33.03 \\
        \textbf{BeliefMem}  & \textbf{42.38}  & \textbf{36.30}  & \textbf{59.88} & \textbf{29.56} \\
        \bottomrule
    \end{tabular}
    \vspace{-3mm}
\end{wraptable}

\subsection{Ablation Studies}
We conduct comprehensive ablation studies to investigate probabilistic memory, belief-aware retrieval, and memory update operations (\emph{Add} and \emph{Merge}) on LoCoMo and ALFWorld (Table~\ref{tab:ablation}). 
As shown, replacing probabilistic memory with standard deterministic memory (w/o belief-based memory) results in clear performance drops in both benchmarks, highlighting the necessity of retaining uncertainty under partial observability. 
Removing belief-aware retrieval eliminates access to memory uncertainty, forcing the agent to discard candidate probabilities at retrieval and consequently degrading performance on both benchmarks.
Furthermore, ablating the update mechanisms undermines BeliefMem's capabilities: removing the \emph{Add} operation prevents the incorporation of new attributes of latent states into the memory bank, while removing \emph{Merge} disables the probability updates over evidence for existing attributes. 
Without them, the memory bank of BeliefMem remains static, notably decreasing the performance arising from dynamic memory updates.
Overall, these results show that each part of our method is vital for achieving reliable memory under partial observability. The full results and detailed analyses are provided in Appendix~\ref{app:full_ablation}.

\subsection{Analysis and Discussion}
\label{sub:further}
\begin{figure}[t]
    \centering
    \includegraphics[width=1\linewidth]{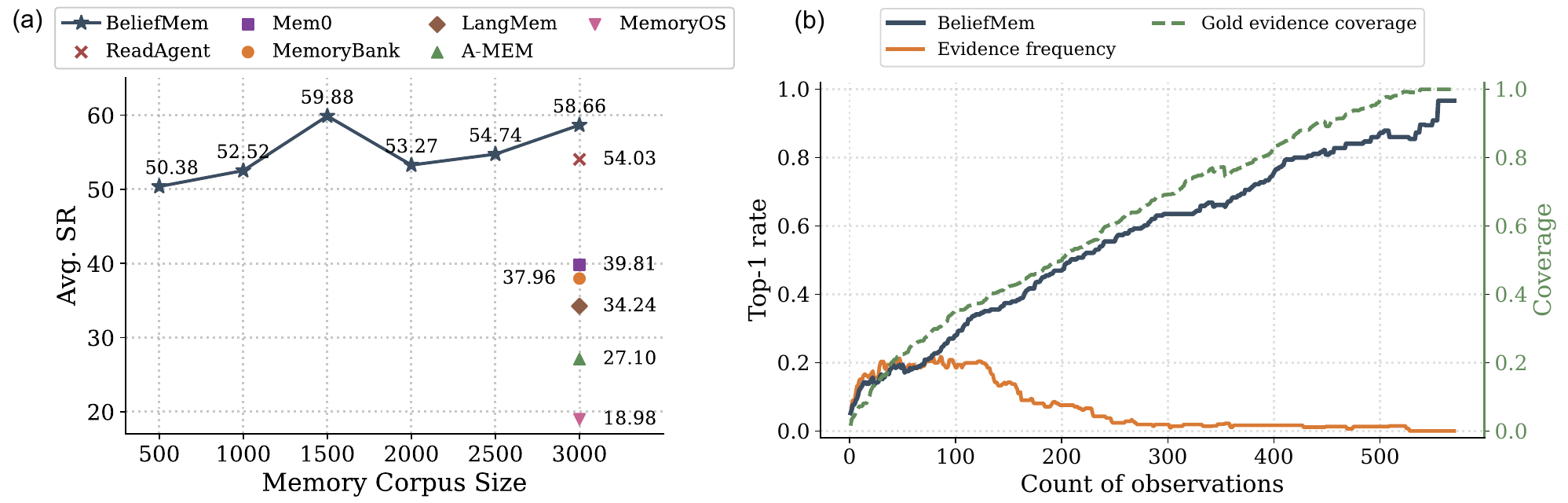}
    \caption{(a) BeliefMem maintains competitive performance across varying memory corpus sizes on ALFWorld, outperforming all baselines with only 50\% of memory corpus. (b) BeliefMem's candidate probabilities reliably converge to the true conclusion as evidence accumulates on LoCoMo, whereas naive frequency-based estimation fails to converge under noisy observations.}
    \label{fig:analysis}
\end{figure}

\textbf{BeliefMem scales robustly under limited memory data.} Figure~\ref{fig:analysis} evaluates BeliefMem on ALFWorld with memory corpus sizes ranging from 500 to 3,000. When using just 50\% of the corpus, BeliefMem already outperforms all baselines, and even with 500 samples, it still surpasses 5 out of 6 baselines.
Beyond this advantage, we observe a trade-off within BeliefMem: using the full memory corpus produces the best performance on seen tasks, whereas a 50\% subset results in superior generalization on unseen tasks. 
This trade-off arises because a richer set of in-distribution memories can bias the agent toward memorizing seen trajectories, at the cost of out-of-distribution generalizability.
These analyses demonstrate that BeliefMem's probabilistic memory enables effective knowledge retention even when data is highly limited. Full results are in Appendix~\ref{app:scaleofalfworld}.

\textbf{BeliefMem achieves reliable belief convergence.} 
To validate whether BeliefMem enables candidate probabilities to converge to the ground truth, we report the Top-1 rate, defined as the proportion of instances where the true conclusion attains the highest confidence among all candidates. In Figure~\ref{fig:analysis}, Top-1 rate of BeliefMem steadily increases as evidence accumulates, achieving 87.68\% of cases where the true conclusion receives the highest probability. 
In contrast, a baseline using raw evidence frequency as confidence fails to converge reliably, as noisy observations distort the frequency of evidence. 
Accordingly, these results demonstrate that BeliefMem's memory update effectively filters noise and raises the confidence of true conclusions over time. Details are provided in Appendix~\ref{app:coverage}.

\begin{wrapfigure}[12]{r}{0.4\textwidth}
\centering
\vspace{-4mm}
\includegraphics[width=\linewidth]{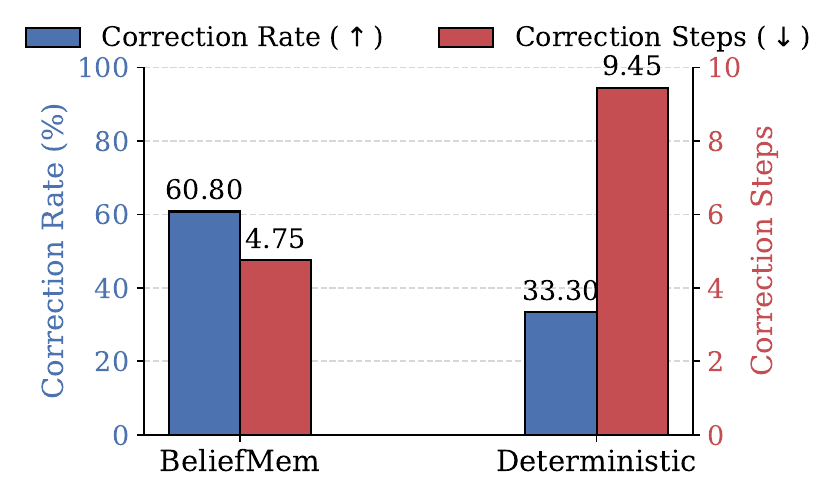}
\caption{BeliefMem vs. deterministic memory under adversarial setting on ALFWorld. }
\label{fig:adv}
\end{wrapfigure}

\textbf{BeliefMem shows strong memory correction in adversarial settings.} We conduct adversarial experiments on ALFWorld benchmark by injecting strongly flawed memory conclusions into the memory bank and observing the correction process (see Appendix~\ref{app:adv} for detailed pipeline). As shown in Figure~\ref{fig:adv}, after updates with valid and noisy observations, BeliefMem achieves a correction rate nearly twice that of the deterministic memory baseline. Furthermore, it achieves this correction notably faster, requiring an average of only 4.75 steps. These results highlight BeliefMem's robustness and stability when handling flawed memories under noisy observations.

\textbf{Hyperparameter analysis.} We evaluate the impact of the retrieval size $K$ and decay rate $\lambda$ on BeliefMem (Table~\ref{tab:sensitivity_hyperparams} in Appendix~\ref{app:furtherresult}). Performance on ALFWorld scales positively with $K$ up to an optimal $K=20$. 
Beyond this ($K=30$), BeliefMem suffers a trade-off: although its SR on seen tasks reaches the best, the SR on unseen tasks drops by 5.22\% as broader retrieval may surface noisy, in-distribution memories that hinder generalization. 
Additionally, variations in $\lambda$ affect the performance of BeliefMem, highlighting the critical role of the decay mechanism in controlling the agent's reliance on early memory, thereby striking a crucial balance between efficient in-distribution exploitation and robust out-of-distribution generalization.

\section{Conclusion}
In this work, we identify a key drawback of prior memory methods in partially observable environments: their deterministic paradigm of storing categorical conclusions inferred from observations results in self-reinforcing error.
To address this issue, we propose BeliefMem, which reframes memory as an approximation of the environment's belief state.
Specifically, BeliefMem maintains multiple candidate conclusions with probabilities for each attribute of the evolving environment, updated via noisy-OR evidence merge as new observations arrive. During retrieval, these probabilistic conclusions enable the agent to reason under uncertainty and select reliable actions toward task goals.
Experiments on the LoCoMo and ALFWorld benchmarks show that our method outperforms well-known baselines on average across diverse scenarios.
Additionally, various analyses in our work illustrate our method's promising capabilities in memory correction and data efficiency. 
Overall, our work introduces a novel perspective on agent memory in partially observable environments and demonstrates its empirical benefits under various settings.

\bibliographystyle{plainnat}
\bibliography{references}


\appendix

\section{More Implementation Details of BeliefMem.}

\subsection{More Details about Memory Update}
\label{app:add}
Given a new observation, the agent first extracts a set of candidate conclusions using the prompt shown in Figure~\ref{fig:memory_prompt}. Each candidate is represented as a structured memory object, including its normalized conclusion, semantic slots, evidence references, temporal information, and belief scores. In practice, an attribute $c$ is formed from stable semantic slots such as subject, predicate, entities, and qualifiers; a candidate conclusion $h$ is the normalized conclusion text/object for that attribute. 
Notably, the extracted \texttt{prob} field is used as the evidence strength $\Delta(o_{t+1}, h)$ in Eq.~\ref{eq:merge}, which measures how strongly the new observation support the conclusion $h$.
We use this value as an LLM-extracted confidence, not as a calibrated posterior probability. For \emph{Add}, we clip this extracted value to $[p_{\min}, p_{\max}]$. Throughout the implementation, the stored \texttt{prob} values are confidence scores used for ranking and updating, not calibrated probabilities.

After extraction, BeliefMem updates the existing memory bank through several operations. If the candidate describes a new conclusion that is not covered by the current memory bank, BeliefMem applies \emph{Add} and inserts it as a new memory entry. If the candidate provides compatible evidence for an existing conclusion (using keyword matching over attribute conclusions), BeliefMem uses \emph{Merge}: the new evidence is attached to the existing memory, and the truth belief is updated with the noisy-OR evidence aggregation in Eq.~\ref{eq:merge}.

\subsection{Contradictory Memory}
\label{app:contra}
For any candidate conclusion $h$, if the observation $o_{t+1}$ provides evidence to support a contradictory conclusion, the current belief of $h$ is reduced to $0.25$, called \emph{Version}. And the previous value is retained as a historical version.
Specifically, we use a rule-based criterion to identify contradictory conclusions: 
Formally, let $(c,h)$ denote an existing memory conclusion of attribute $c$ and $(c,h')$ denote a newly extracted candidate from $o_{t+1}$ via the operation in Appendix~\ref{app:add}. When $h \neq h'$, for same attribute $c$, the new candidate is treated as a contradictory conclusion for $h$.

\subsection{Hyperparameter Configuration}
\label{app:hyper}
BeliefMem uses the same \emph{Add} bounds in Eq.~\ref{eq:add} across both benchmarks. The initial probability interval is $[p_{\min},\, p_{\max}] = [0.7,\, 0.9]$. The decay rate $\lambda$ in Eq.~\ref{eq:decay} is set to $0.5$ for LoCoMo and $0.1$ for ALFWorld.

For LoCoMo, we use random seed $20260413$. Top-$K=20$ for single-hop questions and top-$K=30$ for multi-hop, temporal, and open-domain questions. 
For ALFWorld, we follow the official evaluation, which sets chunk size $512$, query source \texttt{objective}, Contriever retrieval, memory top-$K=20$, and a maximum of $50$ environment steps. The action model is run with seed $42$, temperature $0.0$, top-$p=1.0$, and a maximum generation length of $32$.

$\mathrm{sim}(\cdot)$ in Eq.~\ref{eq:decay} employs a hybrid design. Specifically, it is computed as a linear combination of embedding cosine similarity and lexical overlap (both attribute and evidence), with weights of 0.7 and 0.3, respectively, across all tasks.
In addition, to reduce cost and latency in BeliefMem, we set the maximum number of candidate conclusions per attribute to 4 during retrieval.

\subsection{Memory Costs}
\label{app:cost}
\begin{wrapfigure}{R}{0.5\textwidth}
    \centering
    \includegraphics[width=\linewidth]{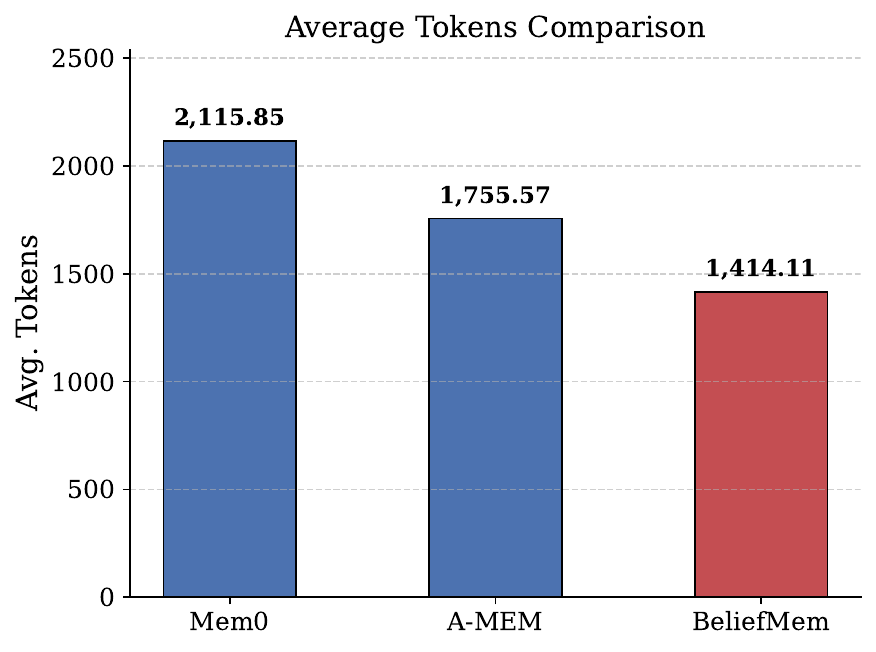}
    \caption{Average token consumption of BeliefMem and competitive baselines on LoCoMo using GPT-4o-mini for each generation.}
    \label{fig:cost}
\end{wrapfigure}
BeliefMem stores one candidate set for each active attribute and optionally preserves historical versions after \emph{Merge}. If attribute $c$ has $M_c$ active candidates and $v_c$ retained versions, memory storage is $O(\sum_c M_c v_c)$ textual entries plus embeddings. 
Updating an observed attribute costs $O(M_c)$ after LLM extraction, since only candidates under the matched attribute are updated. 
Retrieval first scores attributes by semantic similarity and decay, then serializes the top-$K$ attributes and their active candidates, so token cost grows with the number of retrieved candidates rather than only with $K$. 

To reduce this cost, we cap the number of retrieved candidates per attribute. Figure~\ref{fig:cost} reports the average token consumption per generation on LoCoMo. Our method uses fewer tokens than the competitive baselines, confirming that the strategy effectively limits overhead.

\subsection{Hardware and Software}
All base models and benchmarks used in this work are publicly accessible. All experiments were conducted using NVIDIA A800-80GB GPUs with Python $3.11$ and PyTorch $2.4.1$.

\section{Further Experiment Setup}
\subsection{ALFWorld Evaluation Details}
\label{app:alfworld}

\paragraph{Evaluation split.}
For all methods in Section~\ref{sub:mainresult}, we conduct experiments on the official ALFWorld~\citep{shridhar2020alfworld}, evaluating the full 140 episodes of the in-distribution \emph{Seen} set and the full 134 episodes of the out-of-distribution \emph{Unseen} set. 
Both splits cover the six household goal templates (Pick \& Place, Examine in Light, Clean \& Place, Heat \& Place, Cool \& Place, and Pick Two \& Place), and every episode runs under the standard 50-step environment horizon.

\paragraph{Memory bank construction and retrieval.}
We follow the ALFWorld pipeline of~\citet{zhang2026memskill} for all memory methods in this paper. Expert trajectories are collected from the official training split, where each trajectory records the full sequence of observations, actions, and outcomes produced by the demonstrating agent, and are then grouped by task type. 
For every task type, a random subset of training trajectories is sampled as the experience corpus used for memory construction. Evaluation uses the official Seen/Unseen episodes described above, so no evaluation trace is used during memory construction.
We fix the total bank size at 3,000 expert trajectories, distributed across the six task types unless otherwise stated, such as BeliefMem* and the data-size analysis. 
Specifically, the memory bank is constructed once before evaluation begins, with each trajectory written through the native write operation of each baseline method.
At test time, every method, including BeliefMem and all baselines, retrieves up to 20 memory items per observation (Top-$K=20$). The sampled corpus and evaluation episodes are kept identical across methods.

\subsection{LoCoMo Evaluation Details}
For LoCoMo, we follow the official setting in~\citet{maharana2024evaluating}. All baseline methods are reproduced with their open-source settings described in their papers. Additionally, we also present the performance of Mem0 and A-MEM reported in their papers in Section~\ref{sub:mainresult} for clarification. They are denoted with $^{*}$.

\subsection{Preserving Historical Beliefs}
Instead of overwriting an existing belief $p_t$ during \emph{Merge}, BeliefMem retains $p_t^{(c)}$ as an independent historical entry alongside the newly updated current version $p_{t+1}$. This update mechanism is essential for handling temporal queries. While recent observations naturally update the agent's current belief of the environment, queries targeting specific past contexts necessitate access to historical states.

This temporal awareness is achieved through timestamp management. The updated entry $p_{t+1}$ receives the latest timestamp, while the old entry $p_t$ retains its original timestamp. Following the decay mechanism in Eq.~\ref{eq:decay}, the current version is naturally prioritized during default retrieval due to its recency. Simultaneously, the historical version remains fully accessible for queries explicitly referencing earlier time steps, ensuring comprehensive temporal grounding without information loss.

\subsection{Belief Coverage Analysis Experiment Setup}
\label{app:coverage}
To ensure reproducibility and clarity, we detail the setup of belief convergence in Section~\ref{sub:further}.
We conduct our evaluation on the multi-hop task of LoCoMo benchmark, maintaining all hyperparameters at their defaults. 
Specifically, we choose the gold-standard answers of 211 selected samples that can be mapped to a single attribute-level conclusion as the target true states. 
The observations are sampled from the memory corpus associated with these questions, ensuring they contain the necessary evidence to support these ground-truth conclusions. 
Through this rigorous configuration, we effectively validate the capacity of BeliefMem to make the memory belief converge toward the true conclusion.

\subsection{Detailed Pipeline for Adversarial Memory Correction}
\label{app:adv}
To provide further clarity on the adversarial experiments discussed in Section~\ref{sub:further}, this section details the complete experimental pipeline, including adversarial sample construction, update procedures, and evaluation metrics. 

\textbf{Experimental Setup.} The correction process is evaluated through the following steps:
\begin{itemize}
    \item \textbf{Flawed Memory Injection:} We scan the BeliefMem memory bank evaluated on the ALFWorld benchmark to identify strongly flawed conclusions. A memory entry is selected as an adversarial sample if it meets three criteria: (1) it contradicts the optimal action, (2) it is highly ranked (retrieved in the Top-$K$), and (3) the correct conclusion is entirely excluded from the Top-$K$. This strict filtering yields 102 adversarial samples.
    \item \textbf{Observation Generation:} For each sample, we construct a sequence of observations to simulate the update process. We generate 5 \textit{valid observations} based on the correct actions, providing sparse, ground-truth hints. Simultaneously, we construct 5 \textit{noisy observations} derived from incorrect candidate conclusions (strictly excluding the correct one) to serve as adversarial perturbations during the memory update phase.
    \item \textbf{Update Protocol:} BeliefMem is updated using the default settings specified in Appendix~\ref{app:hyper}. As a baseline, the deterministic memory method only stores and updates a single conclusion per sample, as autonomously determined by the agent. Additionally, we run 10 update steps, where each step randomly includes one of the valid or the noisy observation.
\end{itemize}

\textbf{Evaluation Metrics.} We assess memory correction performance using two primary metrics:
\begin{itemize}
    \item \textbf{Correction Rate:} The proportion of samples where the correct conclusion successfully outranks the injected flawed conclusion during retrieval after the update process.
    \item \textbf{Correction Steps:} The average number of update steps required for the correct conclusion to achieve a stably higher retrieval ranking than the flawed conclusion.
\end{itemize}

\section{Further Empirical Results}
\label{app:furtherresult}

\subsection{Full results of BeliefMem on ALFWorld with different memory corpus size.}
\label{app:scaleofalfworld}
In this section, we provide the full results of Figure~\ref{fig:analysis}, as shown in Table~\ref{tab:dataset_size_ablation}.
As detailed, we observe a generalization trade-off related to memory corpus size. 
Specifically, BeliefMem achieves its highest out-of-distribution (ALF-Unseen) success rate of 61.19\% and optimal average performance of 59.88\% using only 1,500 samples, representing exactly 50\% of the sampled memory corpus. 
Additionally, the agent also exhibits maximum behavioral efficiency in novel environments, requiring a minimum of only 29.34 steps to complete tasks.

Conversely, scaling the memory corpus to the full 3,000 samples maximizes in-distribution (ALF-Seen) performance, reaching a peak success rate of 63.57\% with the fewest interaction steps (27.49). 
However, this data increase results in a sharp 7.44\% decline in unseen success rates compared to the 1,500-sample performance. 
This divergence suggests that, in this setting, excessive environment-specific data may induce corpus size overfitting, biasing the agent toward memorizing seen trajectories at the expense of generalizability. We treat this explanation as plausible rather than conclusive, since no additional controlled test of this hypothesis is performed.
Furthermore, BeliefMem demonstrates exceptional low-data robustness; with merely 500 samples (16.67\% of the data), it maintains an average success rate of 50.38\%. 
Overall, these results show that BeliefMem efficiently distills actionable, generalizable memories from highly limited interactions, whereas simply increasing the memory corpus does not monotonically improve its robust environmental understanding.

\begin{table}[t]
\centering
\caption{The performance of BeliefMem on the ALFWorld dataset with varying corpus sizes.}
\label{tab:dataset_size_ablation}
\begin{tabular}{cccccc}
\toprule
\multirow{2}{*}{Corpus Size} & \multicolumn{2}{c}{ALF-Seen} & \multicolumn{2}{c}{ALF-Unseen} & Avg.\ \\
\cmidrule(lr){2-3} \cmidrule(lr){4-5} \cmidrule(lr){6-6}
 & SR $\uparrow$ & \#Steps $\downarrow$ & SR $\uparrow$ & \#Steps $\downarrow$ & SR $\uparrow$ \\
\midrule
500  & 50.00 & 32.51 & 50.75 & 33.80 & 50.38 \\
1000 & 54.29 & 32.78 & 50.75 & 32.62 & 52.52 \\
1500 & 58.57 & 29.77 & \textbf{61.19} & \textbf{29.34} & \textbf{59.88} \\
2000 & 54.29 & 31.09 & 52.24 & 31.56 & 53.27 \\
2500 & 55.00 & 30.60 & 54.48 & 31.81 & 54.74 \\
3000 & \textbf{63.57} & \textbf{27.49} & 53.75 & 30.49 & 58.66 \\
\bottomrule
\end{tabular}
\end{table}

\subsection{Full Results of Ablation Studies}
\label{app:full_ablation}
The complete results of the ablation on ALFWorld and LoCoMo are provided in Tables~\ref{tab:full_ab_alfworld} and~\ref{tab:fullabl_locomo}, respectively.

\paragraph{w/o belief-based memory.}
In this setting, we collapse the probabilistic memory to a deterministic one. 
Specifically, for each attribute, only the single most likely conclusion is kept and retrievable. 
As shown, success rate on ALFWorld drops from 59.88 to 28.71, and average F1 on LoCoMo falls from 42.38 to 22.58. 
Without belief representation over conclusions, the agent acts on overconfident, often incorrect, memories and loses the ability to reason under partial observability.

\paragraph{w/o belief-aware retrieval.}
All candidate conclusions for an attribute are still stored, but their probabilities are discarded during retrieval, making them appear equally likely. 
As observed, the performance drop is more moderate, where ALFWorld success rate declines to 51.77 (a drop of 8.11 absolute SR points compared to full BeliefMem) and LoCoMo average F1 decreases to 28.50.
This indicates that merely retaining multiple hypotheses already preserves a useful degree of uncertainty. 
However, these results change on the more challenging LoCoMo sub‑tasks: on multi‑hop and open‑domain questions, F1 decreases from 40.51 to 27.12 and from 28.73 to 15.89, respectively. 
In these settings, the agent must overcome conflicting evidence, and without probabilities it is unable to judge between competing claims, leading to ambiguous retrieval.

\paragraph{w/o \emph{Add}.}
When \emph{Add} is entirely removed, no new memory is inferred from observations. 
This destroys the dynamic memory update in our method. As a result, the ALFWorld success rate collapses to 22.58\%, and LoCoMo F1 drops to 14.48\%, showing that correct attribution of new evidence is a crucial condition for memory to remain organized and usable.

\paragraph{w/o \emph{Merge}.}
Removing \emph{Merge} means that every new observation creates a separate attribute entry rather than updating an existing one with accumulated evidence. 
Consequently, probabilities are never refined by subsequent observations and remain frozen at their initial values.
ALFWorld success rate falls to 40.81\% and LoCoMo F1 drops to 20.38\%, as the memory stays static and cannot integrate sequential information.

\begin{table}[t]
\centering
\caption{Full results of ablation studies on ALFWorld benchmark (memory corpus size = 1500).}
\label{tab:full_ab_alfworld}
\begin{tabular}{lccc}
\toprule
\multirow{2}{*}{Method} & ALF-Seen & ALF-Unseen & Average \\
 & (SR / \#Steps$\downarrow$) & (SR / \#Steps$\downarrow$) & (SR / \#Steps$\downarrow$) \\
\midrule
w/o belief-based memory    & 27.65 / 39.44 & 29.76 / 38.29 & 28.71 / 38.87 \\
w/o belief-aware retrieval & 54.29 / 31.59 & 49.25 / 33.54 & 51.77 / 32.57 \\
w/o \emph{Add}                    & 19.87 / 43.24 & 25.29 / 39.82 & 22.58 / 41.53 \\
w/o \emph{Merge}                  & 39.29 / 32.38 & 42.32 / 33.67 & 40.81 / 33.03 \\
\textbf{BeliefMem (Ours)}           & \textbf{58.57 / 29.77} & \textbf{61.19 / 29.34} & \textbf{59.88 / 29.56} \\
\bottomrule
\end{tabular}
\end{table}

\begin{table}[t]
\centering
\caption{Full results of ablation studies on LoCoMo benchmark using GPT-4o-mini.}
\label{tab:fullabl_locomo}
\resizebox{\textwidth}{!}{
\begin{tabular}{lccccc}
\toprule
\multirow{2}{*}{Method} & Multi-Hop & Temporal & Open Domain & Single Hop & AVG. \\
 & (F1 / BLEU) & (F1 / BLEU) & (F1 / BLEU) & (F1 / BLEU) & (F1 / BLEU) \\
\midrule
w/o belief-based memory    & 22.45 / 16.57 & 16.82 / 12.34 & 10.11 / 8.76  & 40.93 / 30.02 & 22.58 / 16.92 \\
w/o belief-aware retrieval & 27.12 / 18.98 & 25.67 / 15.43 & 15.89 / 10.12 & 45.31 / 31.55 & 28.50 / 19.02 \\
w/o \emph{Add}                   & 14.56 / 10.21 &  9.14 / 6.88  &  8.73 / 6.54  & 25.48 / 16.91 & 14.48 / 10.14 \\
w/o \emph{Merge}                  & 16.84 / 12.63 & 17.39 / 10.17 & 11.05 / 7.99  & 36.22 / 22.46 & 20.38 / 13.31 \\
BeliefMem (Ours)           & \textbf{40.51 / 32.24} & \textbf{51.88 / 45.78} & \textbf{28.73 / 25.12} & \textbf{48.41 / 42.07} & \textbf{42.38 / 36.30} \\
\bottomrule
\end{tabular}
}
\end{table}

\subsection{Results of hyperparameter analysis}

\textbf{Top-$K$ analysis.} Table \ref{tab:sensitivity_hyperparams} presents a sensitivity analysis of the retrieval size $K$ on ALFWorld, evaluating its impact on both task success rate (SR) and interaction efficiency (Steps). The empirical results demonstrate a clear non-linear relationship between memory retrieval scale and the agent's generalization capabilities, identifying $K=20$ as the optimal threshold for BeliefMem. Specifically, at $K=20$, the model achieves the best generalization and overall efficiency, achieving an average SR of 59.88\% while minimizing the average step to 29.55 steps. When the retrieval size is overly restricted ($K \le 10$), the agent exhibits degraded performance across all metrics. This indicates that an insufficient $K$ fails to retrieve adequate contextual memory.
Conversely, expanding the retrieval size to $K=30$ exposes a generalization trade-off. While a larger memory context maximizes the SR on in-distribution tasks (Seen SR peaks at 61.43\%), it severely compromises out-of-distribution reasoning. The Unseen SR experiences a sharp 8.5\% absolute degradation (from 61.19\% down to 55.97\%), with a corresponding degradation in execution efficiency (Unseen steps increase to 32.43). This structural divergence suggests a trade-off: excessive retrieval surfaces redundant, task-specific noisy memories from seen environments. Rather than augmenting the belief state, these extraneous in-distribution memories may act as noise, impairing the agent's generalization in unseen environments.

\textbf{Decay rate $\lambda$.} To investigate the impact of the hyperparameter decay rate $\lambda$ on both task efficacy and generalization, we conduct a comprehensive sensitivity analysis on ALFWorld. As illustrated in Table~\ref{tab:sensitivity_hyperparams}, removing this term (w/o decay) yields the highest in-distribution SR (63.57\%) but results in the poorest out-of-distribution performance (55.97\% unseen SR) with the longest unseen trajectory length, clearly indicating strong reliance on earlier memories from seen environments. 
However, $\lambda = 0.1$ drastically shifts this dynamic, achieving the best unseen success rate (61.19\%) while sacrificing seen performance. As $\lambda$ is incrementally increased towards 0.9, BeliefMem exhibits a steady recovery in seen environments while maintaining robust unseen generalization, achieving one of the highest average success rates.
Furthermore, the step metrics reveal that higher values of $\lambda$ ($\ge 0.9$) consistently induce more efficient decision-making.
Specifically, the agent uses fewer steps, driving the average trajectory length down to its minimum of 29.00 steps at $\lambda = 1.0$, as the agent can retrieve more related early memories.
Consequently, the choice of $\lambda$ explicitly dictates the agent's reliance on its historical memory. Increasing $\lambda$ encourages the recall of past experiences to achieve optimal in-distribution execution, whereas decreasing $\lambda$ prevents the model from being constrained by prior patterns, strictly benefiting out-of-distribution performance.

\begin{table}[t]
  \centering
  \caption{Sensitivity analysis of hyperparameters (Top-$K$ and $\lambda$) on ALFWorld.}
  \label{tab:sensitivity_hyperparams}
  \small
  \begin{tabular}{lcccccc}
    \toprule
    \multirow{2}{*}{Hyperparameter} & \multicolumn{2}{c}{ALF-Seen} & \multicolumn{2}{c}{ALF-Unseen} & \multicolumn{2}{c}{Avg.} \\
    \cmidrule(lr){2-3} \cmidrule(lr){4-5} \cmidrule(lr){6-7}
       & SR$\uparrow$ & \#Steps$\downarrow$ & SR$\uparrow$ & \#Steps$\downarrow$ & SR$\uparrow$ & Step$\downarrow$ \\
    \midrule
    \multicolumn{7}{l}{\textit{Sensitivity to Top-$K$}} \\
    \midrule
    5  & 50.71 & 33.37 & 53.73 & 32.18 & 52.22 & 32.78 \\
    10 & 50.00 & 32.60 & 52.24 & 32.60 & 51.12 & 32.60 \\
    20 & 58.57 & 29.77 & 61.19 & 29.34 & 59.88 & 29.55 \\
    30 & 61.43 & 29.45 & 55.97 & 32.43 & 58.70 & 30.94 \\
    
    \midrule
    \multicolumn{7}{l}{\textit{Sensitivity to $\lambda$}} \\
    \midrule
    w/o decay & 63.57 & 28.42 & 55.97 & 31.72 & 59.77 & 30.07 \\
    0.1 & 58.57 & 29.77 & 61.19 & 29.34 & 59.88 & 29.55 \\
    0.3 & 59.29 & 28.84 & 58.21 & 29.95 & 58.75 & 29.40 \\
    0.5 & 59.29 & 28.90 & 58.96 & 30.01 & 59.13 & 29.46 \\
    0.7 & 60.00 & 29.06 & 58.21 & 30.13 & 59.10 & 29.60 \\
    0.9 & 62.86 & 27.60 & 58.21 & 30.72 & 60.53 & 29.16 \\
    1.0 & 62.14 & 27.81 & 58.96 & 30.19 & 60.55 & 29.00 \\
    \bottomrule
  \end{tabular}
\end{table}

\begin{figure}[t]
\begin{promptbox}
You are extracting structured long-term memory from one interaction session.

Return JSON only. Do not include markdown fences.
Return a JSON object with a single key "memories" whose value is a list of memory objects.

Return between 14 and 20 memory objects whenever the session contains enough facts.
Prioritize atomic memories that help answer future QA questions.

Each memory object must contain:
- type: one of observation, event, profile, anchor, episode
- canonical_text
- subject
- predicate
- object
- participants
- entities
- qualifiers
- dialog_ids
- time_text
- relative_time
- prob

Rules:
- Only use facts supported by the provided session.
- Keep each memory atomic: one memory should capture one fact, event, or stable profile attribute.
- Prefer profile nodes for stable attributes and preferences.
- Prefer event nodes for time-bounded actions or episodes.
- Prefer observation nodes for one-off factual statements grounded in 1-2 turns.
- Use episode only for a very short episode label, not a long summary paragraph.
- Use exact dialog ids for evidence whenever possible.
- Every returned memory must cite dialog ids from the current session only.
- canonical_text should be concise and specific, usually one sentence or a short clause.
- object should be short when possible.
- time_text must be a short time phrase only, such as "last Saturday", "2022", or "7 May 2023". Never copy an entire sentence into time_text.
- If no time phrase is available, leave time_text empty.
- Do not invent unsupported details.
- Include memories for:
  - important plans, goals, identity, relationship status, jobs, preferences
  - concrete events, activities, meetings, races, support groups, volunteer work
  - key artifacts or entities that future questions may mention
- Capture proper nouns exactly when they appear, including countries, cities, book titles, bands, workshops, conferences, parks, museums, and artists.
- Capture background/profile facts even when mentioned indirectly, such as home country, roots, family structure, children, marriage, support network, books read, favorite activities, places visited, and recurring hobbies.
- For list-like evidence, prefer one memory per distinct item when that will help later QA, such as one book title, one camping location, one music artist, one event, or one activity per memory.
- Favor exact, evidence-grounded wording over paraphrastic summaries.

Session date: {session_date}
Conversation:
{conversation}
\end{promptbox}
\caption{The prompt used for attribute extraction. It restricts the model to output format, fact-based JSON objects grounded in the provided conversation.}
\label{fig:memory_prompt}
\end{figure}

\section{Limitations and Future Work}
While BeliefMem successfully shifts the memory paradigm from storing deterministic conclusions to maintaining a belief representation of the underlying true states, achieving promising performance across diverse scenarios, several limitations remain to be addressed in future work:
\begin{itemize}
    \item \textbf{Lack of theoretical guarantees for belief approximation.} BeliefMem maintains the probability of each candidate conclusion via noisy‑OR evidence aggregation rather than a complete normalized posterior distribution, because exact belief maintenance over an open‑ended hypothesis space is computationally infeasible. Although this approximation provides no formal convergence guarantees, the experimental results in Figure~\ref{fig:analysis} show that as evidence accumulates the candidate probabilities reliably converge toward the true conclusion, demonstrating that the approximation is effective in practice.

    \item \textbf{LLM-extracted evidence strength.} The evidence strength $\Delta$ used in the noisy‑OR update is extracted by LLMs instead of being derived from a calibrated observation likelihood model. This can introduce noise when the model’s confidence estimates are inaccurate. However, the adversarial correction experiments in Section~\ref{sub:further} indicate that BeliefMem is robust to noisy observations, achieving a correction rate for flawed memory entries that is nearly twice that of the deterministic baseline.

    \item \textbf{Computational overhead.} Although we leverage an approximated belief representation, the computational cost remains non-trivial compared to standard deterministic memory baselines, especially during memory writing and merging. Given that our method uses fewer tokens than competitive baselines (Table~\ref{app:cost}), exploring more cost-effective architectures for maintaining and updating belief-based memory represents a promising direction for future work.
\end{itemize}

\section{LLM Usage Statement}
In this paper, we employed the commercial large language model GPT‑5-Chat for language refinement and manuscript polishing. It was not used for generating research ideas, designing methods, or conducting a literature search and discovery.


\end{document}